\documentclass[pmlr]{jmlr}
\usepackage{pgfplots}
\usepgfplotslibrary{groupplots}
\pgfplotsset{compat=1.18}
\usepackage{longtable}
\usepackage{booktabs}
\usepackage{adjustbox}
\usepackage{graphicx}

\makeatletter
\let\Ginclude@graphics\@org@Ginclude@graphics 
\makeatother

\jmlrvolume{260}
\jmlryear{2024}
\jmlrworkshop{ACML 2024}
\jmlrpages{-}
\renewcommand{\thepage}{}
\title[MIAs on Time-Series Models]{Membership Inference Attacks Against \\Time-Series Models}

\author{%
  \Name{Noam Koren} \Email{noam.koren1@ibm.com}\\
  \Name{Abigail Goldsteen} \Email{abigailt@il.ibm.com}\\
  \Name{Guy Amit} \Email{guy.amit@ibm.com}\\
  \Name{Ariel Farkash} \Email{arielf@il.ibm.com}\\
  \addr IBM Research, Haifa, Israel
}

\editors{Vu Nguyen and Hsuan-Tien Lin}

\begin{document}

\maketitle

\begin{abstract}
Analyzing time-series data that contains personal information, particularly in the medical field, presents serious privacy concerns. Sensitive health data from patients is often used to train machine learning models for diagnostics and ongoing care. Assessing the privacy risk of such models is crucial to making knowledgeable decisions on whether to use a model in production or share it with third parties. Membership Inference Attacks (MIA) are a key method for this kind of evaluation, however time-series prediction models have not been thoroughly studied in this context.
We explore existing MIA techniques on time-series models, and introduce new features, focusing on the seasonality and trend components of the data. Seasonality is estimated using a multivariate Fourier transform, and a low-degree polynomial is used to approximate trends. We applied these techniques to various types of time-series models, using datasets from the health domain. Our results demonstrate that these new features enhance the effectiveness of MIAs in identifying membership, improving the understanding of privacy risks in medical data applications.

\end{abstract}


\section{Introduction}
There is a clear conflict between the ever-increasing interest in analyzing personal data to enhance and improve processes, and the need to preserve the privacy of data subjects. 
In the medical domain, sensitive data from patients is often used to train machine learning (ML) models that aid physicians in diagnostics and treatment. ML models are also utilized within medical devices and applications to predict malfunctions and improve ongoing care. 

Assessing the privacy risk of such models is crucial to enable making knowledgeable decisions on whether to use a model in production, share it with third parties, or deploy it in patients' homes. Privacy risk assessment is often achieved by running membership inference attacks against the models and measuring their success rate. 

Membership inference attacks (MIA) attempt to distinguish between samples that were part of a target model’s training data (called members) and samples that were not (non-members), based on the model's outputs. Many such attacks are based on training a binary classifier as an attack model \cite{shokri}. These attacks can be applied to various model types, including classification \cite{shokri}, regression \cite{attack_reg}, graph \cite{attack_graph}, and generative \cite{attack_gen} models. However, MIA against time-series prediction models, has not yet been properly researched.

This paper addresses this gap by evaluating existing membership inference approaches on time-series forecasting models and introducing new features specifically designed for these models. Our main contribution is the addition of two novel features that exploit the trend and seasonality components of time-series data. The trend is approximated by fitting a low-degree polynomial and the seasonality is estimated using the Discrete Fourier Transform (DFT). 

Since time series fundamentally consist of trend and seasonality components, it is reasonable to assume that time-series models are more adept at accurately estimating these elements in series encountered during training.
Additionally, various state-of-the-art forecasting models, such as Neural Fourier Transform (NFT) \cite{nft}, TimesNet \cite{wu2022timesnet}, etc., 
specifically incorporate those components into their design. Consequently, when targeting a time-series prediction model, there is a significant likelihood that the model will precisely estimate the series' seasonality and trend of its training data, providing a strategic advantage in MIAs. This underscores the importance of considering these features when assessing the vulnerability of time-series models.

The impact of adding seasonality and trend as input features to MIA models is empirically evaluated by testing different combinations of existing and new features. The evaluation is performed on six time-series prediction models, using two medical datasets. The results demonstrate significant improvements across multiple prediction horizons, ranging from 3\% to 26\%, confirming the efficacy of the proposed attack features. This is an important first step towards proper privacy assessment methods for time-series models, which have so far been mostly overlooked.

\section{Background}
\label{related}
\subsection{Time-Series Forecasting models}

Time series forecasting has evolved significantly, initially relying on linear models like ARIMA \cite{arima} and Exponential Smoothing \cite{exponential}. However, with deep learning advancements, neural network architectures such as LSTM \cite{yu2019review} and GRU \cite{dey2017gate} showed superior performance over traditional methods.
 
Recently, Convolutional Neural Networks (CNNs) \cite{CNN} and Temporal Convolutional Networks (TCNs) \cite{tcn} have demonstrated state-of-the-art results. The Transformer architecture \cite{transformers_survay} was also adapted for forecasting, with models like AutoFormer \cite{autoformer} and FEDformer \cite{fedformer} as leading architectures. 
However, \cite{dlinear} proposed DLinear, a simple linear model, challenging the efficacy of transformers. The subsequent models TimesNet \cite{wu2022timesnet}, and PatchTST \cite{patchtst} improved upon DLinear, while the NFT model \cite{nft} emerged as a top-performing multivariate time-series model.


\textbf{Multidimensional Fourier Transform.} The Fourier Transform has been widely used in time series analysis to identify periodic patterns or cycles in the data \cite{yi2023survey}. By converting time-series data into the frequency domain, one can identify the main frequencies at which these cycles occur \cite{nussbaumer1982fast}.

Several forecasting models use Fourier Transforms for better performance. Autoformer employs Fast Fourier Transform for autocorrelation \cite{autoformer}, FEDformer focuses on key frequencies \cite{fedformer}, and the Fourier Neural Operator approximates partial differential equations operators with Fourier Transforms \cite{fno}. TimesNet uses Fourier Transforms for feature decomposition to capture periodic patterns \cite{wu2022timesnet}.

The \emph{Multidimensional Fourier Transform (MFT)} \cite{tolimieri2012mathematics} extends the traditional Fourier Transform to handle multi-dimensional data. We drew inspiration from the Neural Fourier Transform (NFT) \cite{nft}, and used a 2-dimensional Discrete Fourier Transform (DFT) to extract the seasonality of the time series. 

\subsection{Membership Inference Attacks}
Membership inference attacks (MIAs) represent a significant privacy threat in machine learning. In these attacks, an adversary aims to determine whether a specific data record, $x$, was included in the training set of a model, $D$. 
If successful, the attack can reveal sensitive information about individuals, such as their medical history, financial status, or personal preferences. Moreover, MIAs can also be used to identify individuals who are part of a specific group or community, potentially leading to discrimination, or physical harm. 

Formally, given access to a machine learning model $M$, the attacker seeks to ascertain the membership of a data sample $x$ in $D$, i.e; to check if $x \in D$. To this end the attacker typically analyzes $M$'s outputs, and produces numeric characteristics (features), that will enable it to distinguish members of the training data from none members.
Such features may include $M$'s loss~\cite{shokri} on the sample, the log-probabilities~\cite{log_probs} and entropy of the outputs.

In the context of time-series forecasting models, a malicious attacker seeks to determine whether a specific time series was utilized in the model's training dataset, such as a patient's ECG test results. This type of attack poses a significant threat in industries like healthcare and finance, where sensitive time-series data is frequently leveraged to develop predictive models, and the unauthorized disclosure of such information could have severe consequences.


\section{Related work}
In the realm of time series, Hisamoto et al. studied membership inference on sequence-to-sequence (seq2seq) models in the context of machine translation, where the output is a chained sequence of classifications \cite{s2s}. 
This differs from medical sequence modeling whose input features and outputs are numerical and continuous. 

Pyrgelis et al. \cite{pyrgelis2017knock} presented the first study on the feasibility of MIAs on aggregate location time series, modeling the problem as a classification task to distinguish whether a target user is part of an aggregate. 
Their empirical evaluation on mobility datasets shows that MIAs are a privacy threat, influenced by the adversary’s prior knowledge, data characteristics, number of users, and aggregation timeframe. 

Similarly, Voyez et al. \cite{voyez2022} explored the vulnerability of aggregated time-series data to MIAs, introducing a linear programming-based attack that leverages the correlation between the length of the published time series and the size of the aggregated data. 
Their experiments demonstrate that aggregated time series data can be highly susceptible to privacy breaches, emphasizing the need for better privacy-preserving techniques, particularly in the medical domain.

However, to our knowledge, risk assessment in general and MIA specifically has not been thoroughly explored on ML models trained on numerical time-series data.
This presented us with an opportunity for novel applications and advancements in attacking time-series models, potentially unlocking new insights and methodologies in this area.

\section{Methodology}
\label{method}
\subsection{Problem Statement}
This study focuses on MIA on multivariate time-series forecasting models. We assume that the attacker can access a complete sample that was either used in model training or not. 

In time-series data, training samples consist of data points up to time $T$ (lookback), denoted as $\mathbf{y} = [y_{1}, \ldots, y_{T}] \in \mathbb{R}^{M \times T}$, and the model predicts $H$ data points onward (horizon), denoted as $\mathbf{Y} = [y_{T+1}, \ldots, y_{T+H}] \in \mathbb{R}^{M \times H}$, where $y_t \in \mathbb{R}^M$ for $t = 1, \ldots, T+H$, and $M$ is the number of variables.

To simplify, we consider a lookback window of length $t \leq T$, ending at the most recent observation $y_T$. This window serves as the input (sample) to the model and is denoted $\mathbf{X} \in \mathbb{R}^{M \times t} = [y_{T-t+1}, \ldots, y_T]$. The forecast of $\mathbf{Y}$ is represented as $\hat{\mathbf{Y}}$.

Our task is to determine if a specific sample, $\mathbf{X}$, is part of the training data, $\mathbf{D}$, i.e; $\mathbf{X\in D}$ by comparing the real future values, $\mathbf{Y}$, and the models predicted values, $\hat{\mathbf{Y}}$.


\subsection{Features for MIA}
\label{Features}
In the context of MIA, the attack features are the set of attributes or characteristics that the attack model leverages to determine whether a given data sample was a part of the training set (member) or not (non-member). As in any ML model, selecting the correct attack features is critical, as they form the basis upon which the attack model makes its predictions. An optimal set of attack features can significantly improve the success of the attack, potentially posing a much higher privacy risk.

\begin{figure}[ht]
  \centering
  \includegraphics[width=\linewidth]{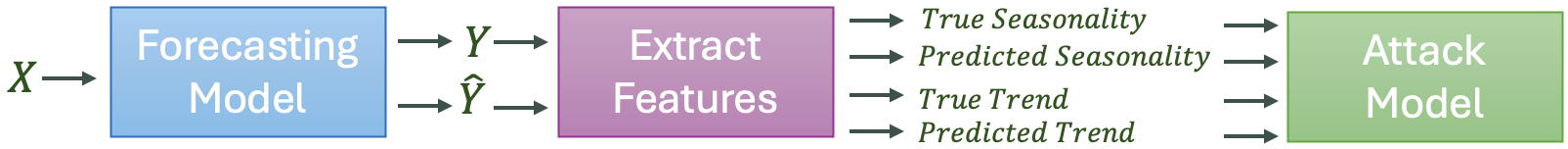}
  \caption{Flowchart illustrating the process from the initial forecasting model to the final attack model, highlighting how extracted features are used for MIAs.}
  \label{fig:model_process}
\end{figure}

Our goal is to find attack features that will yield good MIA results for time-series models. This involves leveraging the characteristics of time-series data by identifying features that capture its unique aspects. Hence, the introduced features isolate the seasonality and trend components of the model’s prediction, contrasted with those of the true data. 
Figure \ref{fig:model_process} illustrates this process from the initial forecasting model to the final attack setup.

Given that time-series data inherently includes components such as trend and seasonality, models trained on such data are particularly good at capturing those elements. This proficiency is leveraged by a variety of forecasting models, including Neural Fourier Transform (NFT) \cite{nft}, TimesNet \cite{wu2022timesnet}, Fedformer \cite{fedformer}, Autoformer \cite{autoformer}, Fourier Neural Operator (FNO) \cite{fno}, N-BEATS \cite{nbeats}, NeuralProphet \cite{triebe2021neuralprophet}, ARIMA \cite{arima}, etc., all of which explicitly integrate these elements into their predictions. Thus, when attacking a time-series prediction model, we aim to take advantage of the model's ability to accurately predict the series' trend and seasonality, thus offering a tactical advantage in MIAs. This highlights the critical need to consider these characteristics when evaluating the susceptibility of time-series models to such privacy threats.

In this work, to effectively capture the seasonality, we employ the Multidimensional Fourier Transform, which excels in extracting periodic patterns from time-series data \cite{musbah2019sarima}.
We identify the predominant trend through a low-degree polynomial fit, allowing us to find the principal direction while filtering variations \cite{masry1996multivariate}.

\subsubsection{Seasonality}
We detect seasonality in multivariate temporal data with the 2-dimensional Discrete Fourier Transform (2D-DFT) as done in \cite{nft}. This method breaks down the dataset into its frequency components, considering both the range of variables and the timeline. This approach is essential for datasets where the interaction between different variables can create new seasonality patterns that are not seen in the univariate context.

The 2D-DFT is applied to the matrix $\mathbf{Y} \in \mathbb{R}^{M \times H}$, where \(M\) is the number of variables, and \(H\) is the number of predicted time points. This process involves two sequential 1D-DFTs. First, a column-wise 1D-DFT is applied to \(Y\) using the Fourier matrix \(F_M\), which captures transformations across the variables. Next, a row-wise 1D-DFT is performed using the Fourier matrix \(F_H\), which encodes the temporal structure of the data.

The 2D-DFT can be compactly represented as: \begin{equation} \label{eq
Coeffs mat} \mathbf{C} = \mathbf{F_M Y F_H}^\top \end{equation}

Here, the matrix \(C\) contains the Fourier coefficients.





Following are the Fourier matrices \( F_M\) and \( F_H \) that achieve the desired Fourier transformation:

\newcommand{\cdotsTwo}{\mathinner{\cdot\cdot}}
\newcommand{\vdotsTwo}{\mathinner{\vbox{\baselineskip4pt \lineskiplimit0pt
\hbox{.}\hbox{.}}}}

\[
F_M= 
\begin{bmatrix}
\cos(2\pi \cdot 0 \cdot \frac{0}{M}) & \cdotsTwo & \cos(2\pi \cdot 0 \cdot \frac{M-1}{M}) \\
\vdotsTwo & \vdotsTwo & \vdotsTwo \\
\cos(2\pi \cdot \frac{M}{2} \cdot \frac{0}{M}) & \cdotsTwo & \cos(2\pi \cdot \frac{M}{2} \cdot \frac{M-1}{M}) \\
\sin(2\pi \cdot 0 \cdot \frac{0}{M}) & \cdotsTwo & \sin(2\pi \cdot 0 \cdot \frac{M-1}{M}) \\
\vdotsTwo & \vdotsTwo & \vdotsTwo \\
\sin(2\pi \cdot \frac{M}{2} \cdot \frac{0}{M}) & \cdotsTwo & \sin(2\pi \cdot \frac{M}{2} \cdot \frac{M-1}{M}) \\
\end{bmatrix}
\]
\[
F_H= 
\begin{bmatrix}
\cos(2\pi \cdot 0 \cdot \frac{0}{H}) & \cdotsTwo & \cos(2\pi \cdot 0 \cdot \frac{H-1}{H}) \\
\vdotsTwo & \vdotsTwo & \vdotsTwo \\
\cos(2\pi \cdot \frac{H}{2} \cdot \frac{0}{H}) & \cdotsTwo & \cos(2\pi \cdot \frac{H}{2} \cdot \frac{H-1}{H}) \\
\sin(2\pi \cdot 0 \cdot \frac{0}{H}) & \cdotsTwo & \sin(2\pi \cdot 0 \cdot \frac{H-1}{H}) \\
\vdotsTwo & \vdotsTwo & \vdotsTwo \\
\sin(2\pi \cdot \frac{H}{2} \cdot \frac{0}{H}) & \cdotsTwo & \sin(2\pi \cdot \frac{H}{2} \cdot \frac{H-1}{H}) \\
\end{bmatrix}    
\]



The input features to the attack derived from this method are:
\begin{enumerate}
    \item Coefficients of the Fourier series corresponding to the true values: \[\mathbf{C} = \mathbf{F_1}^\top \times \mathbf{Y} \times \mathbf{F_2} \]
    \item Coefficients of the Fourier series corresponding to the model’s predicted values: \[\mathbf{\hat{C}} = \mathbf{F_1}^\top \times \mathbf{\hat{Y}} \times \mathbf{F_2} \]
    \item The \( L_2 \) norm between the coefficients of the true and predicted values: 
    \[ ||\mathbf{C}-\mathbf{\hat{C}}||_2 \]
\end{enumerate}



\subsubsection{Trend}
Consider a multivariate time series \( Y\) with \( H \) time points and \( M \) variables. Each variable's series is approximated using a polynomial of degree \( d \).
This approximation can be represented as: 
\begin{equation}
\mathbf{Y = P \times A}
\end{equation}
where \(A\) is the coefficients matrix, and \( P \) is the Vandermonde matrix, constructed from the time vector \( t = \frac{[0, 1, \dots, H-1]}{H} \). \( P \) contains powers of \( t \) up to \( d-1 \), has dimensions \( d \times H \) and is defined as:
\[
P = \begin{bmatrix}
    1 & 1 & \cdots & 1 \\
    t_1 & t_2 & \cdots & t_H \\
    t_1^2 & t_2^2 & \cdots & t_H^2 \\
    \vdots & \vdots & \ddots & \vdots \\
    t_1^{d-1} & t_2^{d-1} & \cdots & t_H^{d-1}
\end{bmatrix}
\]
where \( t_i = \frac{i-1}{H} \) for \( i = 1, 2, \ldots, H \).

The coefficients matrix \(A\) is obtained by the least squares solution: 
$\mathbf{A} = (\mathbf{P}^T \mathbf{P})^{-1} \mathbf{P}^T \mathbf{Y}$

The input features to the attack derived from this method are:
\begin{enumerate}
    \item Coefficients of the polynomial outlining the trend of the true values: \[ \mathbf{A} = (\mathbf{P}^T \mathbf{P})^{-1} \mathbf{P}^T \mathbf{Y} \]
    \item Coefficients of the polynomial outlining the trend of the model’s predicted values:\[ \mathbf{\hat{A}} = (\mathbf{P}^T \mathbf{P})^{-1} \mathbf{P}^T \mathbf{\hat{Y}} \]
    \item The \( L_2 \) norm between the coefficients of the true and predicted values: 
    \[ ||\mathbf{A}-\mathbf{\hat{A}}||_2 \]
\end{enumerate}

\subsubsection{Mean Absolute Scaled Error and Mean Squared Error}

Additionally, this study investigates incorporating the Mean Absolute Scaled Error (MASE), a scaled measure for assessing forecast accuracy by comparing the mean absolute error of a model against a naïve baseline forecast, as outlined by  \cite{hyndman2006another}, and the Mean Squared Error (MSE), \cite{das2004mean}, metrics as features for the attack model. 


\begin{align*}
\text{MASE} &= \frac{\frac{1}{H} \sum_{i=1}^{H} |y_{T+i} - \hat{y}_{T+i}|}{\frac{1}{H-1} \sum_{i=2}^{H} |y_{T+i} - y_{T+i-1}|}
\end{align*}

\begin{align*} 
\text{MSE} &= \frac{1}{H} \sum_{i=1}^{H} (y_{T+i} - \hat{y}_{T+i})^2
\end{align*}

These metrics are extended to the multivariate case by averaging the respective univariate values across all variables.

\section{Experimental Setup}
\label{eval}

\subsection{Threat Model and MIA Attack Setup}


Following previous privacy assessment studies~\cite{framework, sok, anderson2024my}, we adopt a gray-box threat model, assuming the attacker has access to both a subset of the model's training data and a set of non-training samples.
The access to this data, allows estimating a worst case privacy risk for a model before deployment, without the need of developing shadow models~\cite{shokri}.


To execute the attack, we leverage the privacy risk assessment framework from~\cite{framework}, which builds upon recent breakthroughs in MIAs~\cite{carlini2022membership, shokri}. 
This framework leverages the ensemble approach, creating many specialized attack models for different subsets of the data. The framework harnesses a diverse set of input features extracted from the target model's inputs and outputs. Through an exhaustive grid search, it systematically explores various attack model architectures, hyperparameters, and preprocessing techniques to identify the optimal configuration that yields maximum attack performance.



The attack models trained by the risk assessment framework utilize various combinations of the following features: (1) Seasonality denoted as, \textbf{\textit{S}}, (2) Trend denoted as, \textbf{\textit{T}} (with a polynomial degree of 4 like done in \cite{nft, nbeats}), (3) \textbf{\textit{MASE}}, (4) \textbf{\textit{MSE}}, and (5) Predicted Values denoted as, \textbf{\textit{PV}}.


In this evaluation, we applied the framework to conduct five attack instances, each with three runs. 
An instance refers to executing the entire attack optimization process on a different random data sample, while the runs involve different splits of the data sample to fit and infer the attack model within each instance. 
For each instance, a sample of 450 members and 450 non-members was chosen at random. Results were averaged across all runs and instances to ensure robustness.




\subsection{Datasets}
In this evaluation, two multivariate time-series medical datasets were used:
\begin{itemize}
\item \textbf{EEG}: 36-lead EEG database, which contains more than 1000 EEG recordings dating from 2002 to the present, sampled at a frequency of 250 Hz \cite{obeid2016temple}. Our subset includes data from 32 patients, and the first 3-leads for each patient.
\item \textbf{ECG}: Georgia 12-Lead ECG Challenge Database, curated by Emory University \cite{Goldberger2000}. The complete database contains ECG recordings of over 10,000 individuals, sampled at a frequency of 500 Hz, and collected from various healthcare settings worldwide. Our subset features ECG time-series data from 600 individuals.
\end{itemize}

Data preprocessing included outlier removal using the Interquartile Range method, imputation of missing values via mean substitution, and data standardization.

For both datasets, The data was partitioned into three distinct subsets: 42.5\% of the patients were used for training the model, 15\% for validation, and the remaining 42.5\% were reserved as non-member data points for the attack model. The validation set was used to tune the models parameters, thus creating strong models to attack. The non-member data points remained uninvolved in training or validating the models but were used in subsequent attack experiments. Additionally, the data was split into lookbacks and horizons using the sliding window approach.
Statistics on the datasets can be found in Table \ref{fig:datasets}.

\begin{table*}[h!]
\centering
\caption{Datasets Statistics}
\begin{tabular}{lccccc}
 \toprule
Dataset & Num of Variables & Timesteps & Lookback & Prediction Horizons \\
\midrule
EEG & 3 & 9620519 & 100 & 1, 5, 10, 15, 20 \\
ECG   & 12 & 2393563 & 100 & 1, 5, 10, 15, 20, 25, 30 \\ \bottomrule
\end{tabular}
\label{fig:datasets}
\end{table*}

\subsection{Models}
We performed attacks against various state-of-the-art time-series forecasting architectures:

\begin{itemize}
    \item \textbf{DLinear}: Featured a dimension of 16 and a dropout rate of 0.1 \cite{dlinear}.
    \item \textbf{Temporal Convolutional Network (TCN)}: Configured with channels set to [2, 2], a kernel size of 2, and a dropout rate of 0.2 \cite{tcn}.
    \item \textbf{Long Short Term Memory (LSTM)}: Featured a 2-layer structure with hidden dimensions set to 50 \cite{yu2019review}.
    \item \textbf{Neural Fourier Transform (NFT)}: Configured with Fourier granularity of 8 for the seasonality blocks and a polynomial degree of 4 for the trend blocks, comprising 2 blocks per stack \cite{nft}.
    \item \textbf{TimesNet}: Model dimension was set to 16 and a dropout rate of 0.1 \cite{wu2022timesnet}.
    \item \textbf{PatchTST}: This transformer model included one encoder and decoder layer, a model dimension of 16, and a dropout rate of 0.1 \cite{patchtst}.
\end{itemize}

\begin{figure*}[htbp]
\centering
\begin{adjustbox}{center}
\begin{tikzpicture}
  \begin{groupplot}[
    group style={
      group size=2 by 1, 
      horizontal sep=3cm,
      vertical sep=1.8cm, 
    },
    width=6cm,
    height=3cm,
    legend style={at={(-0.3,-0.7)}, anchor=north, legend columns=-1}
  ]

  \pgfplotsset{
    nft/.style={blue, densely dashdotted},
    tcn/.style={teal, densely dashed},
    lstm/.style={purple, densely dashdotted},
    timesnet/.style={brown, densely dashdotted},
    patchtst/.style={green, dashdotted},
    dlinear/.style={orange, dashdotted},
  }
    
    \nextgroupplot[xlabel={Horizon}, ylabel={MSE}, title={EEG}]
    \addplot[nft] coordinates {(1,0.06) (5, 0.09) (10, 0.1) (15, 0.1) (20, 0.11)};
    \addplot[timesnet] coordinates {(1,0.04) (5, 0.07) (10, 0.1) (15, 0.1) (20, 0.11)};
    \addplot[patchtst] coordinates {(1,0.03) (5, 0.06) (10, 0.09) (15, 0.1) (20, 0.11)};
    \addplot[dlinear] coordinates {(1,0.03) (5, 0.05) (10, 0.08) (15, 0.09) (20, 0.10)};
    \addplot[tcn] coordinates {(1,0.06) (5, 0.09) (10, 0.1) (15, 0.11) (20, 0.12)};
    \addplot[lstm] coordinates {(1,0.06) (5, 0.09) (10, 0.1) (15, 0.11) (20, 0.12)};
    
    \nextgroupplot[xlabel={Horizon}, ylabel={MSE}, title={ECG}]
    \addplot[nft] coordinates {(1,0.02) (5,0.08) (10,0.08) (15,0.12) (20,0.15) (25,0.15) (30,0.21)};
    \addplot[timesnet] coordinates {  (1,0.02) (5,0.08) (10,0.08) (15,0.1) (20,0.15) (25,0.15) (30,0.17)};
    \addplot[patchtst] coordinates {  (1,0.02) (5,0.05) (10,0.08) (15,0.1) (20,0.15) (25,0.15) (30,0.18)};
    \addplot[dlinear] coordinates { (1,0.02) (5,0.06) (10,0.08) (15,0.11) (20,0.14) (25,0.16) (30,0.19)};
    \addplot[tcn] coordinates {(1,0.03) (5,0.07) (10,0.1) (15,0.12) (20,0.16) (25,0.18) (30,0.21) };
    \addplot[lstm] coordinates {(1,0.21) (5,0.24) (10,0.41) (15,0.42) (20,0.43) (25,0.39) (30,0.49) };

  \addlegendentry{NFT} 
  \addlegendentry{TimesNet}
  \addlegendentry{PatchTST}
  \addlegendentry{DLinear}
  \addlegendentry{TCN}
  \addlegendentry{LSTM}
  \end{groupplot}
\end{tikzpicture}
\end{adjustbox}
\caption{MSE values of models with varying horizons on ECG and EEG datasets}
\label{fig:mse_results}
\end{figure*}
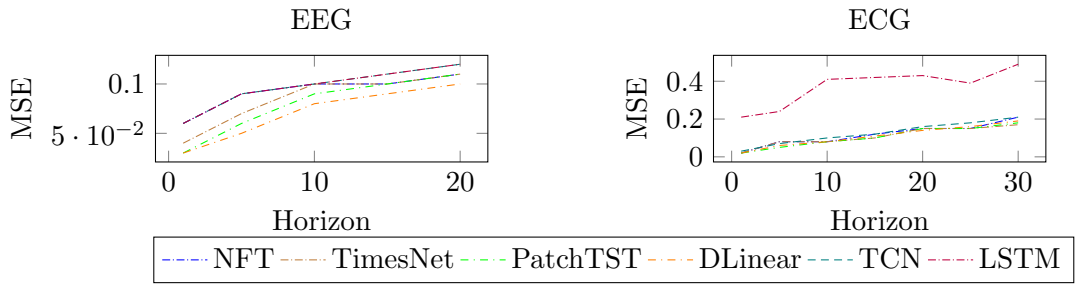

The parameters for each model were chosen based on their performance on a validation set, ensuring optimal configuration for our analysis.
In Table \ref{tab:params}, the number of parameters for each model is detailed. For all models, the MSE loss was utilized during model training. Figure \ref{fig:mse_results} presents the performance (MSE) of each model on the test set (non-members) for different prediction horizons.


\section{Results}
To assess the robustness of the proposed features for MIAs on time-series models we compare them to baseline attacks, selected from standard gray-box approaches. In grey-box, where the attacker has 
to both a subset of the model’s training data and a set of non-training samples, typical attacks are based on loss or predictions, similar to those in black-box mode since in both the attacker relies on loss or prediction information \cite{mia_attacks}. In all experiments, four baselines were employed: loss-based attacks using MSE, MASE, and a combination of both, as well as attacks based on predicted values.

The results were evaluated using two key metrics: Area Under the Receiver Operating Characteristic curve (AUC-ROC) and True Positive Rate (TPR) at a fixed False Positive Rate (FPR) of 1\%. 

\subsection{Area Under the ROC Curve Results}
\subsubsection{Performance Against Loss Based Attacks}

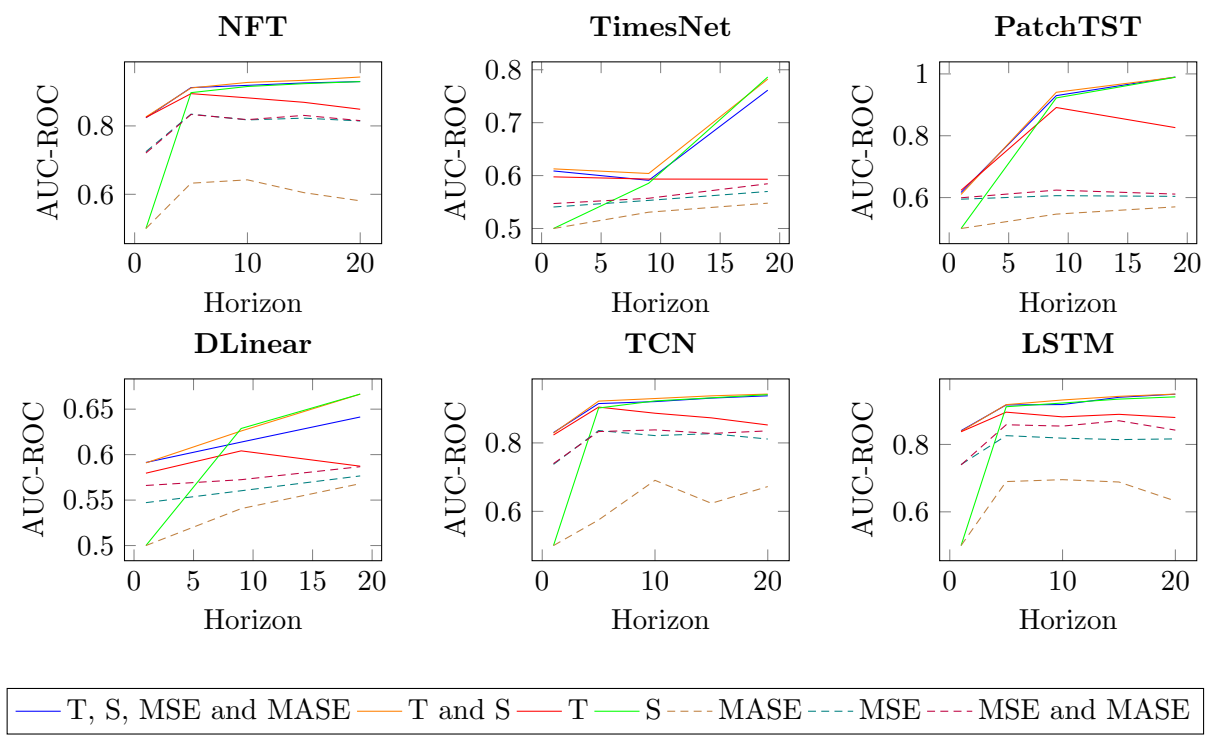
\begin{figure*}[htbp]
\centering
\begin{adjustbox}{center}
\begin{tikzpicture}
  \begin{groupplot}[
    group style={
      group size=3 by 2, 
      horizontal sep=2cm,
      vertical sep=1.8cm, 
    },
    width=5cm,
    height=4cm,
    legend style={at={(-1.3,-0.7)}, anchor=north, legend columns=-1} 
  ]

  \pgfplotsset{
    All/.style={blue, solid},
    TS/.style={orange, solid},
    Trend/.style={red, solid},
    Seasonality/.style={green, solid},
    MASE/.style={brown, densely dashed},
    MSE/.style={teal, densely dashed},
    MSEMASE/.style={purple, densely dashed},
  }

    \nextgroupplot[xlabel={Horizon}, ylabel={{AUC-ROC}}, title={\textbf{NFT}}]
            
    \addplot[All] coordinates {(1, 0.8238345679012345) (5, 0.9120880658436213) (10, 0.9185792592592593) (15, 0.9256729218106996) (20, 0.9295815637860082)};
    \addplot[TS] coordinates {(1, 0.8263572016460905) (5, 0.910956378600823) (10, 0.9271409758965314) (15, 0.9333175308641974) (20, 0.9429991769547325)};
    \addplot[Trend] coordinates {(1, 0.824759341563786) (5, 0.8943937448559671) (10, 0.8820690534979423) (15, 0.8690467489711935) (20, 0.8487367901234567)};
    \addplot[Seasonality] coordinates {(1, 0.5) (5, 0.8972887242798354) (10, 0.9152731687242799) (15, 0.923465843621399) (20, 0.9295647736625515)};
    \addplot[MASE] coordinates {(1, 0.5) (5, 0.6325013991769548) (10, 0.6422786831275721) (15, 0.6048841152263374) (20, 0.5809359670781892)};
    \addplot[MSE] coordinates {(1, 0.7246195061728395) (5, 0.8343910562414266) (10, 0.8175581893004115) (15, 0.8227155555555555) (20, 0.8144133333333332)};
    \addplot[MSEMASE] coordinates {(1, 0.7206979423868314) (5, 0.8334589599700712) (10, 0.8183586008230452) (15, 0.8307367901234569) (20, 0.8154571193415638)};
    \nextgroupplot[xlabel={Horizon}, ylabel={{AUC-ROC}}, title={\textbf{TimesNet}}]
            
    \addplot[All] coordinates {(1, 0.6087486419753085) (9, 0.5910447736625514) (19, 0.7614123456790124)};
    \addplot[TS] coordinates {(1, 0.6127172016460904) (9, 0.6039627983539095) (19, 0.782326255144033)};
    \addplot[Trend] coordinates {(1, 0.597594732510288) (9, 0.5934309465020575) (19, 0.5930906995884773)};
    \addplot[Seasonality] coordinates {(1, 0.5) (9, 0.5855865020576132) (19, 0.7863091358024692)};
    \addplot[MASE] coordinates {(1, 0.5) (9, 0.5308166255144032) (19, 0.5478551440329218)};
    \addplot[MSE] coordinates {(1, 0.5406168724279835) (9, 0.5531107818930041) (19, 0.5699772839506173)};
    \addplot[MSEMASE] coordinates {(1, 0.5471443621399177) (9, 0.5571494650205762) (19, 0.5845818930041152)};
    \nextgroupplot[xlabel={Horizon}, ylabel={{AUC-ROC}}, title={\textbf{PatchTST}}]
            
    \addplot[All] coordinates {(1, 0.6165448559670781) (9, 0.9297581893004114) (19, 0.9899733333333334)};
    \addplot[TS] coordinates {(1, 0.6101779423868312) (9, 0.9405776131687243) (19, 0.9893255967078189)};
    \addplot[Trend] coordinates {(1, 0.6235142386831276) (9, 0.8913078189300411) (19, 0.8262452674897119)};
    \addplot[Seasonality] coordinates {(1, 0.5) (9, 0.922403621399177) (19, 0.9890814814814816)};
    \addplot[MASE] coordinates {(1, 0.5) (9, 0.5464972839506173) (19, 0.5700072427983539)};
    \addplot[MSE] coordinates {(1, 0.594799670781893) (9, 0.606232427983539) (19, 0.6039420576131687)};
    \addplot[MSEMASE] coordinates {(1, 0.5993907818930041) (9, 0.624159670781893) (19, 0.6112801646090534)};
    \nextgroupplot[xlabel={Horizon}, ylabel={{AUC-ROC}}, title={\textbf{DLinear}}]
            
    \addplot[All] coordinates {(1, 0.5913534156378601) (9, 0.6137822222222222) (19, 0.6413674074074074)};
    \addplot[TS] coordinates {(1, 0.5909033744855967) (9, 0.6259886419753087) (19, 0.6665369547325103)};
    \addplot[Trend] coordinates {(1, 0.5795835390946502) (9, 0.6041303703703703) (19, 0.587285267489712)};
    \addplot[Seasonality] coordinates {(1, 0.5) (9, 0.6287293827160494) (19, 0.6665779423868312)};
    \addplot[MASE] coordinates {(1, 0.5) (9, 0.5406232098765432) (19, 0.5680610699588478)};
    \addplot[MSE] coordinates {(1, 0.5471883127572017) (9, 0.5602602469135802) (19, 0.5765700411522634)};
    \addplot[MSEMASE] coordinates {(1, 0.5660791769547324) (9, 0.5724421399176954) (19, 0.5867344855967078)};
    \nextgroupplot[xlabel={Horizon}, ylabel={{AUC-ROC}}, title={\textbf{TCN}}]
            
    \addplot[All] coordinates {(1, 0.8301815089163236) (5, 0.9149700411522634) (10, 0.9207030452674896) (15, 0.9309148971193416) (20, 0.9376676543209875)};
    \addplot[TS] coordinates {(1, 0.8287068312757201) (5, 0.9221647736625513) (10, 0.9297222908093278) (15, 0.9375805761316872) (20, 0.9427208230452674)};
    \addplot[Trend] coordinates {(1, 0.8231961042524005) (5, 0.9046635390946502) (10, 0.8867641700960219) (15, 0.8732141563786009) (20, 0.8523795884773662)};
    \addplot[Seasonality] coordinates {(1, 0.5) (5, 0.9019766255144033) (10, 0.9227479286694104) (15, 0.9317754732510286) (20, 0.9409397530864195)};
    \addplot[MASE] coordinates {(1, 0.5) (5, 0.574879890260631) (10, 0.6905913854595335) (15, 0.6235267489711933) (20, 0.6723893004115227)};
    \addplot[MSE] coordinates {(1, 0.7372211796982169) (5, 0.8361493552812073) (10, 0.8213625788751716) (15, 0.8270474074074073) (20, 0.8112867489711935)};
    \addplot[MSEMASE] coordinates {(1, 0.7395734430727022) (5, 0.8331880658436214) (10, 0.8380663923182443) (15, 0.8276676543209878) (20, 0.8351741563786008)};
    \nextgroupplot[xlabel={Horizon}, ylabel={{AUC-ROC}}, title={\textbf{LSTM}}]
            
    \addplot[All] coordinates {(1, 0.8409122085048011) (5, 0.917150781893004) (10, 0.918349218106996) (15, 0.9398564609053498) (20, 0.9490390123456789)};
    \addplot[TS] coordinates {(1, 0.8389311385459534) (5, 0.9180589300411522) (10, 0.9318834705075445) (15, 0.9426693004115225) (20, 0.9493922633744856)};
    \addplot[Trend] coordinates {(1, 0.8376081755829906) (5, 0.8957557201646089) (10, 0.8817913854595337) (15, 0.8893071604938273) (20, 0.8799331687242798)};
    \addplot[Seasonality] coordinates {(1, 0.5) (5, 0.9118574485596709) (10, 0.9228810425240057) (15, 0.9344697942386831) (20, 0.9411552263374485)};
    \addplot[MASE] coordinates {(1, 0.5) (5, 0.6898290809327847) (10, 0.6954908641975308) (15, 0.6885122633744856) (20, 0.6322200823045268)};
    \addplot[MSE] coordinates {(1, 0.7404193141289438) (5, 0.8260526748971192) (10, 0.818613278463649) (15, 0.8143461728395062) (20, 0.8164146502057614)};
    \addplot[MSEMASE] coordinates {(1, 0.739295144032922) (5, 0.8583990855052583) (10, 0.8543679012345676) (15, 0.8702528395061726) (20, 0.8424651851851852)};
\addlegendentry{T, S, MSE and MASE}
\addlegendentry{T and S}
\addlegendentry{T}
\addlegendentry{S}
\addlegendentry{MASE}
\addlegendentry{MSE}
\addlegendentry{MSE and MASE}
  \end{groupplot}
\end{tikzpicture}
\end{adjustbox}
\caption{AUC-ROC Performance Across Different Models on EEG Data}
\label{fig:EEG}
\end{figure*}

\begin{figure*}[htbp]
\centering
\begin{adjustbox}{center}
\begin{tikzpicture}
  \begin{groupplot}[
    group style={
      group size=3 by 2, 
      horizontal sep=2cm,
      vertical sep=1.8cm, 
    },
    width=5cm,
    height=4cm,
    legend style={at={(-1.3,-0.7)}, anchor=north, legend columns=-1} 
  ]

  \pgfplotsset{
    All/.style={blue, solid},
    TS/.style={orange, solid},
    Trend/.style={red, solid},
    Seasonality/.style={green, solid},
    MASE/.style={brown, densely dashed},
    MSE/.style={teal, densely dashed},
    MSEMASE/.style={purple, densely dashed},
  }

    \nextgroupplot[xlabel={Horizon}, ylabel={{AUC-ROC}}, title={\textbf{NFT}}]
            
    \addplot[All] coordinates {(1, 0.5876335802469137) (5, 0.576959012345679) (10, 0.5772207407407406) (15, 0.5760103703703703) (20, 0.5647343209876543) (30, 0.578892510288066)};
    \addplot[TS] coordinates {(1, 0.5912858436213992) (5, 0.5734925102880658) (10, 0.5785545679012347) (15, 0.5739749794238682) (20, 0.57967670781893) (30, 0.5649224691358025)};
    \addplot[Trend] coordinates {(1, 0.5975827983539095) (5, 0.5879277366255143) (10, 0.5856967901234568) (15, 0.5770846090534979) (20, 0.57568329218107) (30, 0.5738060905349793)};
    \addplot[Seasonality] coordinates {(1, 0.5) (5, 0.5635908289241621) (10, 0.5803330041152264) (15, 0.5655651028806584) (20, 0.579963950617284) (30, 0.5750156378600823)};
    \addplot[MASE] coordinates {(1, 0.5) (5, 0.5289655692729767) (10, 0.5293338271604938) (15, 0.5290725925925925) (20, 0.534641975308642) (30, 0.534525596707819)};
    \addplot[MSE] coordinates {(1, 0.533518683127572) (5, 0.5438948148148148) (10, 0.5461841975308641) (15, 0.5496080658436214) (20, 0.5506732510288066) (30, 0.5673410699588477)};
    \addplot[MSEMASE] coordinates {(1, 0.5317418106995885) (5, 0.5504096296296297) (10, 0.5606637037037037) (15, 0.5479919341563786) (20, 0.5532804938271605) (30, 0.554164279835391)};
    \nextgroupplot[xlabel={Horizon}, ylabel={{AUC-ROC}}, title={\textbf{TimesNet}}]
            
    \addplot[All] coordinates {(1, 0.7192520164609053) (10, 0.9724948148148147) (24, 0.8665367901234567) (30, 0.8665367901234567)};
    \addplot[TS] coordinates {(1, 0.6980120164609054) (10, 0.9728276543209876) (24, 0.8801399176954732) (30, 0.8801399176954732)};
    \addplot[Trend] coordinates {(1, 0.6902921810699588) (10, 0.8722646913580248) (24, 0.6517965432098765) (30, 0.6517965432098765)};
    \addplot[Seasonality] coordinates {(1, 0.5) (10, 0.9663641152263374) (24, 0.8753958847736625) (30, 0.8753958847736625)};
    \addplot[MASE] coordinates {(1, 0.5) (10, 0.6173297119341563) (24, 0.6780569547325104) (30, 0.6780569547325104)};
    \addplot[MSE] coordinates {(1, 0.6657009053497942) (10, 0.6313708641975309) (24, 0.6060177777777778) (30, 0.6060177777777778)};
    \addplot[MSEMASE] coordinates {(1, 0.6575655967078189) (10, 0.670597695473251) (24, 0.7055731687242799) (30, 0.7055731687242799)};
    \nextgroupplot[xlabel={Horizon}, ylabel={{AUC-ROC}}, title={\textbf{PatchTST}}]
            
    \addplot[All] coordinates {(1, 0.8158264197530866) (10, 0.9994976131687243) (14, 0.9990688065843623) (20, 0.9999030452674897) (24, 0.999616049382716) (30, 0.999616049382716)};
    \addplot[TS] coordinates {(1, 0.7225252674897119) (10, 0.9996738271604938) (14, 0.9991321810699588) (20, 0.999929218106996) (24, 0.9995763786008233) (30, 0.9995763786008233)};
    \addplot[Trend] coordinates {(1, 0.720305596707819) (10, 0.9952883950617284) (14, 0.9819657613168724) (20, 0.9813955555555556) (24, 0.9739743209876544) (30, 0.9739743209876544)};
    \addplot[Seasonality] coordinates {(1, 0.5) (10, 0.9996610699588478) (14, 0.9993886419753085) (20, 0.9999448559670782) (24, 0.9998172839506173) (30, 0.9998172839506173)};
    \addplot[MASE] coordinates {(1, 0.5) (10, 0.6420713580246913) (10, 0.6678704526748971) (14, 0.6065922633744856) (20, 0.5884707818930041) (24, 0.7934135802469136) (30, 0.7934135802469136)};
    \addplot[MSE] coordinates {(1, 0.8420110288065843) (10, 0.8434929218106996) (10, 0.8393254320987653) (14, 0.8276668312757203) (20, 0.7912628806584363) (24, 0.7609878189300411) (30, 0.7609878189300411)};
    \addplot[MSEMASE] coordinates {(1, 0.8489238683127572) (10, 0.845499341563786) (10, 0.8470750617283951) (14, 0.8201988477366255) (20, 0.8013282304526749) (24, 0.8887567078189299) (30, 0.8887567078189299)};
    \nextgroupplot[xlabel={Horizon}, ylabel={{AUC-ROC}}, title={\textbf{DLinear}}]
    \addplot[All] coordinates {(1, 0.6107820576131686) (10, 0.6045096296296296) (14, 0.5910357201646091) (20, 0.6246641975308641) (24, 0.598671769547325) (30, 0.598671769547325)};
    \addplot[TS] coordinates {(1, 0.6175027160493827) (10, 0.6228431275720164) (14, 0.5964352263374485) (20, 0.6370503703703704) (24, 0.5953726748971193) (30, 0.5953726748971193)};
    \addplot[Trend] coordinates {(1, 0.6187861728395061) (10, 0.6013293827160494) (14, 0.5876436213991768) (20, 0.6465655967078189) (24, 0.598560329218107) (30, 0.598560329218107)};
    \addplot[Seasonality] coordinates {(1, 0.5) (10, 0.6092893827160493) (14, 0.5837917695473251) (20, 0.6353269135802468) (24, 0.59040987654321) (30, 0.59040987654321)};
    \addplot[MASE] coordinates {(1, 0.5) (10, 0.5240470781893004) (14, 0.5239683950617284) (20, 0.5285965432098766) (24, 0.5334999176954733) (30, 0.5334999176954733)};
    \addplot[MSE] coordinates {(1, 0.5523204938271605) (10, 0.5490581069958849) (14, 0.5560982716049382) (20, 0.5547486419753087) (24, 0.5678897119341564) (30, 0.5678897119341564)};
    \addplot[MSEMASE] coordinates {(1, 0.5436939917695474) (10, 0.5593710288065843) (14, 0.5621922633744856) (20, 0.5614215637860083) (24, 0.5752079012345679) (30, 0.5752079012345679)};
    
    \nextgroupplot[xlabel={Horizon}, ylabel={{AUC-ROC}}, title={\textbf{TCN}}]
            
    \addplot[All] coordinates {(1, 0.5913618106995885) (5, 0.5773657613168723) (10, 0.5796464197530865) (15, 0.5814609053497942) (20, 0.5799753086419753) (25, 0.5702158024691357) (30, 0.5890232098765431)};
    \addplot[TS] coordinates {(1, 0.5891818930041152) (5, 0.5854306172839505) (10, 0.5775227983539094) (15, 0.5922065843621399) (20, 0.5890561316872429) (25, 0.5773823868312757) (30, 0.5878067489711933)};
    \addplot[Trend] coordinates {(1, 0.6061038683127572) (5, 0.566826913580247) (10, 0.5754442798353908) (15, 0.5750424691358025) (20, 0.5803349794238684) (25, 0.571955720164609) (30, 0.5817015637860082)};
    \addplot[Seasonality] coordinates {(1, 0.5) (5, 0.5779228806584362) (10, 0.5759810699588478) (15, 0.5810686419753086) (20, 0.587519670781893) (25, 0.5753601646090536) (30, 0.5688737448559671)};
    \addplot[MASE] coordinates {(1, 0.5) (5, 0.5313628806584362) (10, 0.5261572016460905) (15, 0.5323150617283952) (20, 0.5365920987654321) (25, 0.5381083127572016) (30, 0.540148806584362)};
    \addplot[MSE] coordinates {(1, 0.5354892181069959) (5, 0.54902) (10, 0.5548413168724279) (15, 0.571912098765432) (20, 0.5493519341563786) (25, 0.5565616460905349) (30, 0.5633705349794239)};
    \addplot[MSEMASE] coordinates {(1, 0.535675061728395) (5, 0.5510079012345679) (10, 0.5502739094650205) (15, 0.5511916049382716) (20, 0.5555066666666667) (25, 0.5568928395061727) (30, 0.5522044444444444)};

    \nextgroupplot[xlabel={Horizon}, ylabel={{AUC-ROC}}, title={\textbf{LSTM}}]
            
    \addplot[All] coordinates {(1, 0.5929433744855965) (5, 0.5744241975308642) (10, 0.5571083127572016) (15, 0.5549512757201646) (20, 0.5572391769547325) (25, 0.5766762139917695) (30, 0.5887107818930041)};
    \addplot[TS] coordinates {(1, 0.5951353086419753) (5, 0.5621190672153635) (10, 0.5599723456790124) (15, 0.5662735802469135) (20, 0.5684495473251029) (25, 0.5767061728395061) (30, 0.5787468312757202)};
    \addplot[Trend] coordinates {(1, 0.5929891358024691) (5, 0.5559874897119341) (10, 0.5504995884773664) (15, 0.5484083950617284) (20, 0.5628074074074074) (25, 0.5667710288065844) (30, 0.579362633744856)};
    \addplot[Seasonality] coordinates {(1, 0.5) (5, 0.5612071604938271) (10, 0.560531522633745) (15, 0.5631855144032922) (20, 0.5643986831275719) (25, 0.5638235390946502) (30, 0.5764628806584362)};
    \addplot[MASE] coordinates {(1, 0.5) (5, 0.5348678189300411) (10, 0.5283422222222222) (15, 0.537764609053498) (20, 0.5372498765432099) (25, 0.5284697942386831) (30, 0.5290245267489712)};
    \addplot[MSE] coordinates {(1, 0.5556605761316872) (5, 0.5626395884773663) (10, 0.5528948148148148) (15, 0.5578246913580247) (20, 0.5540737448559672) (25, 0.5636928669410151) (30, 0.5607005761316872)};
    \addplot[MSEMASE] coordinates {(1, 0.5577484773662552) (5, 0.5570795884773662) (10, 0.5511501234567902) (15, 0.5537868312757201) (20, 0.5498697942386831) (25, 0.5678044444444446) (30, 0.5731463374485597)};
\addlegendentry{T, S, MSE and MASE}
\addlegendentry{T and S}
\addlegendentry{T}
\addlegendentry{S}
\addlegendentry{MASE}
\addlegendentry{MSE}
\addlegendentry{MSE and MASE}
  \end{groupplot}
\end{tikzpicture}
\end{adjustbox}
\caption{AUC-ROC Performance Across Different Models on ECG Data}
\label{fig:ECG}
\end{figure*}
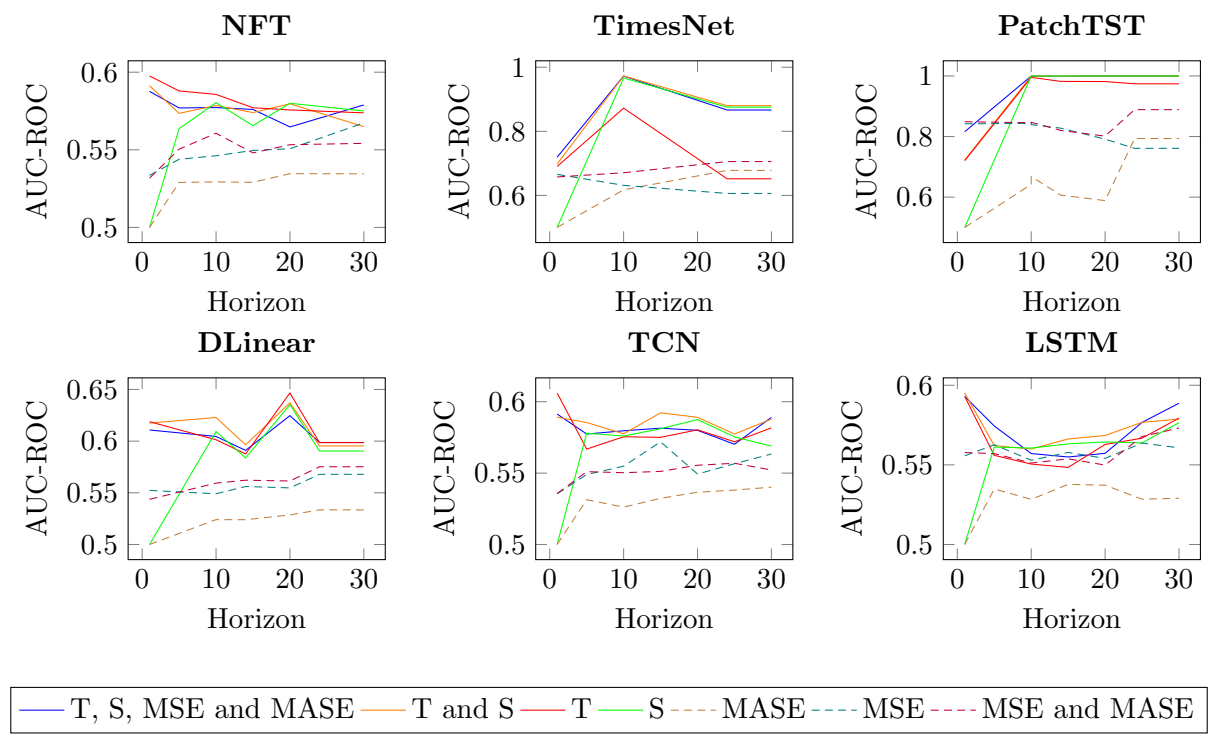

Figures \ref{fig:EEG} and \ref{fig:ECG} highlight the strong performance of the Seasonality (S) and Trend (T) features across various time-series models and datasets. The highest AUC-ROC values were achieved with feature combinations that included the Trend or Seasonality features, outperforming the baseline attacks. 
For the EEG dataset, when looking at the different models, the improvement percentage of the best-performing feature combination compared to the MSE-only attack, averaged across horizons, ranged from \textbf{8.44\%} (with average std \textbf{0.007}) to \textbf{26.41\%} (with average std \textbf{0.006}). For the ECG dataset, improvements ranged from \textbf{2.97\%} (with average std \textbf{0.008}) to \textbf{24.55\%} (with averaged std \textbf{0.007}). The TimesNet model showed the best improvements in both datasets.

The MASE feature achieved the lowest attack performance, however, the combination of MASE and MSE generally surpassed attacks that use the MSE feature alone.

Overall, the results indicate that the Seasonality and Trend features provide a robust solution for membership inference attacks for time-series forecasting models. Its consistent performance across different models, datasets, and horizons highlights its effectiveness.

\textbf{Analysis of Attack Performance by Prediction Horizon.} We analyzed the relative attack AUC-ROC as a function of the prediction horizon. To this end, we computed, for each dataset, the correlation between the prediction horizon (previously denoted by $H$) and the improvement percentage (MSE-only attack relative to the highest attack value). See Table \ref{tab:correlations_improvement} for the resulting correlations. We consistently observed a positive correlation for all the models except for NFT and TCN on the ECG dataset. We assume that the negative correlation can be traced back to the difference in architecture. Notably, the NFT and TCN models both are based on convolutional layers, which capture the trend and seasonality in a different manner. 
The results presented in the table highlight the greater improvement in vulnerability of time series prediction models to attacks as the prediction horizon increases, compared to the MSE baseline attack.
This suggests an enhanced attack vector against these models, which is not present in other model architectures.

\begin{table}[h]
\centering
\caption{Correlation between Prediction Horizons and Improvement Percentages}
\label{tab:correlations_improvement}
\begin{tabular}{|c|c|c|}
\hline
\textbf{Model} & \textbf{Correlation EEG} & \textbf{Correlation ECG} \\
\hline
PatchTST & 0.859  & 0.942  \\
TimesNet & 0.796 & 0.732  \\
DLinear & 0.883 & 0.177  \\
LSTM & 0.482     & 0.376  \\
TCN & 0.780       & -0.150 \\
NFT  & 0.089      & -0.931 \\
\hline
\end{tabular}
\end{table}




\textbf{Analysis of Attack Performance by Model.}
 Table \ref{tab:params} presents the average AUC-ROC results for each model across features and horizons. For the ECG dataset, PatchTST emerged as the most vulnerable model, indicating it is highly susceptible to attacks. In contrast, DLinear was the least vulnerable and hardest to attack. Notably, PatchTST has the highest number of parameters (3,000,000), while DLinear has the fewest (800). In the EEG dataset, DLinear consistently showed superior resilience against attacks, and the vulnerability of the other models varied across the datasets.
 

\begin{table}[h]
\centering
    \caption{Number of Model Parameters VS Average AUC-ROC Results Across Features and Horizons for EEG and ECG Datasets}
    \label{tab:params}
    \begin{tabular}{lccccccc}
        \toprule
Model                & DLINEAR & TCN & LSTM & NFT & TimesNet & PatchTST \\
\midrule
\# Parameters & \(8 \times 10^2\) & \(1 \times 10^3\) & \(34 \times 10^3\) & \(13 \times 10^4\) & \(16 \times 10^4\) & \(3 \times 10^6\)     \\
\midrule
EEG Average AUC-ROC & 0.59 & 0.82 & 0.83 & 0.82 & 0.6 & 0.72 \\
ECG Average AUC-ROC & 0.58 & 0.56 & 0.56 & 0.56 & 0.72 & 0.83 \\
        \bottomrule
    \end{tabular}
\end{table}

\subsubsection{Performance Against Predicted Values Based Attacks}
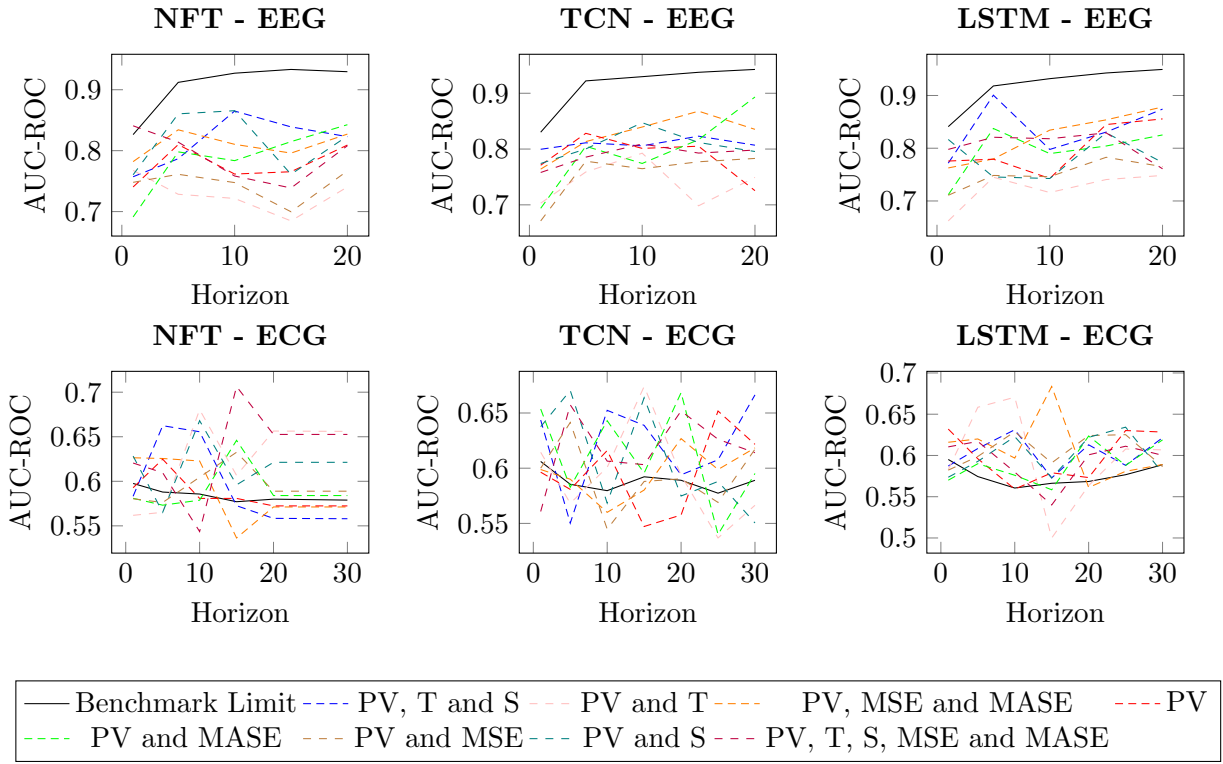
\begin{figure*}[htbp]
\centering
\begin{adjustbox}{center}
\begin{tikzpicture}
  \begin{groupplot}[
    group style={
      group size=3 by 2, 
      horizontal sep=2cm,
      vertical sep=1.8cm, 
    },
    width=5cm,
    height=4cm,
    legend style={at={(-1.2,-0.7)}, anchor=north, legend columns=5} 
  ]

  \pgfplotsset{
    uppbound/.style={black, solid},
    All/.style={blue, densely dashed},
    TS/.style={orange, densely dashed},
    Trend/.style={red, densely dashed},
    Seasonality/.style={green, densely dashed},
    MASE/.style={brown, dashed},
    MSE/.style={teal, dashed},
    MSEMASE/.style={purple, dashed},
    PL/.style={pink, dashed},
  }

    \nextgroupplot[xlabel={Horizon}, ylabel={{AUC-ROC}}, title={\textbf{NFT - EEG}}]
            
    \addplot[uppbound] coordinates {(1, 0.8263572016460905) (5, 0.9120880658436213) (10, 0.9271409758965314) (15, 0.9333175308641974) (20, 0.9295815637860082)};            
    \addplot[All] coordinates {(1, 0.7572000000000001) (5, 0.7868) (10, 0.8650000000000001) (15, 0.839) (20, 0.823)};
    \addplot[PL] coordinates {(1, 0.7676000000000001) (5, 0.7283999999999999) (10, 0.7216000000000001) (15, 0.6847) (20, 0.7404)};
    \addplot[TS] coordinates {(1, 0.782) (5, 0.8342) (10, 0.8104) (15, 0.796) (20, 0.8271999999999999)};
    \addplot[Trend] coordinates {(1, 0.7404) (5, 0.8099999999999999) (10, 0.7615999999999999) (15, 0.7654) (20, 0.8092)};
    \addplot[Seasonality] coordinates {(1, 0.6912) (5, 0.7979999999999999) (10, 0.7836000000000001) (15, 0.8148) (20, 0.8426)};
    \addplot[MASE] coordinates {(1, 0.748) (5, 0.7612) (10, 0.7476) (15, 0.6997) (20, 0.7667999999999999)};
    \addplot[MSE] coordinates {(1, 0.7604) (5, 0.8603999999999999) (10, 0.8656) (15, 0.7611999999999999) (20, 0.8240000000000001)};
    \addplot[MSEMASE] coordinates {(1, 0.8408000000000001) (5, 0.814) (10, 0.7584) (15, 0.7387) (20, 0.8080999999999999)};
    \nextgroupplot[xlabel={Horizon}, ylabel={{AUC-ROC}}, title={\textbf{TCN - EEG}}]
            
    \addplot[uppbound] coordinates {(1, 0.8301815089163236) (5, 0.9221647736625513) (10, 0.9297222908093278) (15, 0.9375805761316872) (20, 0.9427208230452674)};
    \addplot[All] coordinates {(1, 0.7996) (5, 0.8111999999999999) (10, 0.8058) (15, 0.8231000000000002) (20, 0.8069)};
    \addplot[PL] coordinates {(1, 0.7030000000000001) (5, 0.7588999999999999) (10, 0.7926) (15, 0.6983) (20, 0.7492000000000001)};
    \addplot[TS] coordinates {(1, 0.764) (5, 0.8134) (10, 0.8394999999999999) (15, 0.8679) (20, 0.8352999999999999)};
    \addplot[Trend] coordinates {(1, 0.771) (5, 0.8282) (10, 0.8012000000000001) (15, 0.8051) (20, 0.7254)};
    \addplot[Seasonality] coordinates {(1, 0.6937999999999999) (5, 0.805) (10, 0.7742) (15, 0.8168000000000001) (20, 0.8933)};
    \addplot[MASE] coordinates {(1, 0.6712) (5, 0.7782) (10, 0.7647999999999999) (15, 0.7772000000000001) (20, 0.7831999999999999)};
    \addplot[MSE] coordinates {(1, 0.7741) (5, 0.8002) (10, 0.8472) (15, 0.8118000000000001) (20, 0.7955999999999999)};
    \addplot[MSEMASE] coordinates {(1, 0.758) (5, 0.7855) (10, 0.8091999999999999) (15, 0.7928) (20, 0.7982)};
    \nextgroupplot[xlabel={Horizon}, ylabel={{AUC-ROC}}, title={\textbf{LSTM - EEG}}]
            
    \addplot[uppbound] coordinates {(1, 0.8409122085048011) (5, 0.9180589300411522) (10, 0.9318834705075445) (15, 0.9426693004115225) (20, 0.9493922633744856)};
    \addplot[All] coordinates {(1, 0.7717) (5, 0.9006000000000001) (10, 0.7977000000000001) (15, 0.8306) (20, 0.8742000000000001)};
    \addplot[PL] coordinates {(1, 0.6621999999999999) (5, 0.745) (10, 0.716) (15, 0.7406) (20, 0.7484)};
    \addplot[TS] coordinates {(1, 0.7626000000000001) (5, 0.7797) (10, 0.8343) (15, 0.8538000000000001) (20, 0.8781999999999999)};
    \addplot[Trend] coordinates {(1, 0.7757000000000001) (5, 0.7792000000000001) (10, 0.7426999999999999) (15, 0.8449) (20, 0.8555999999999999)};
    \addplot[Seasonality] coordinates {(1, 0.7118) (5, 0.8374) (10, 0.7898) (15, 0.8042) (20, 0.8251999999999999)};
    \addplot[MASE] coordinates {(1, 0.7108) (5, 0.7484) (10, 0.746) (15, 0.7827999999999999) (20, 0.7651000000000001)};
    \addplot[MSE] coordinates {(1, 0.8166) (5, 0.7454999999999999) (10, 0.7426) (15, 0.8291999999999999) (20, 0.7728999999999999)};
    \addplot[MSEMASE] coordinates {(1, 0.7978000000000001) (5, 0.8208) (10, 0.8180000000000001) (15, 0.8291) (20, 0.761)};

    \nextgroupplot[xlabel={Horizon}, ylabel={{AUC-ROC}}, title={\textbf{NFT - ECG}}]
                    
    \addplot[uppbound] coordinates {(1, 0.5975827983539095) (5, 0.5879277366255143) (10, 0.5856967901234568) (15, 0.5770846090534979) (20, 0.579963950617284) (30, 0.578892510288066)};
    \addplot[All] coordinates {(1, 0.5831999999999999) (5, 0.6624) (10, 0.6556) (15, 0.5726) (20, 0.5584) (30, 0.558)};
    \addplot[PL] coordinates {(1, 0.5618000000000001) (5, 0.5652) (10, 0.6803999999999999) (15, 0.6065999999999999) (20, 0.6564000000000001) (30, 0.656)};
    \addplot[TS] coordinates {(1, 0.6264000000000001) (5, 0.6256000000000002) (10, 0.6228) (15, 0.5364) (20, 0.5712) (30, 0.5712)};
    \addplot[Trend] coordinates {(1, 0.5928) (5, 0.6248) (10, 0.5808) (15, 0.5808000000000001) (20, 0.5724) (30, 0.5724)};
    \addplot[Seasonality] coordinates {(1, 0.5809) (5, 0.5731999999999999) (10, 0.5788) (15, 0.6459999999999999) (20, 0.5840000000000001) (30, 0.5840)};
    \addplot[MASE] coordinates {(1, 0.5804) (5, 0.5762) (10, 0.6046) (15, 0.6325999999999999) (20, 0.5889) (30, 0.5889)};
    \addplot[MSE] coordinates {(1, 0.6283000000000001) (5, 0.5651999999999999) (10, 0.6680000000000001) (15, 0.596) (20, 0.6213) (30, 0.6213)};
    \addplot[MSEMASE] coordinates {(1, 0.6205) (5, 0.6096) (10, 0.5436) (15, 0.7064) (20, 0.6526000000000001) (30, 0.6526000000000001)};
    \nextgroupplot[xlabel={Horizon}, ylabel={{AUC-ROC}}, title={\textbf{TCN - ECG}}]
                  
    \addplot[uppbound] coordinates {(1, 0.6061038683127572) (5, 0.5854306172839505) (10, 0.5796464197530865) (15, 0.5922065843621399) (20, 0.5890561316872429) (25, 0.5773823868312757) (30, 0.5890232098765431)};    
    \addplot[All] coordinates {(1, 0.6436) (5, 0.5502) (10, 0.6524) (15, 0.6388) (20, 0.594) (25, 0.6076) (30, 0.6664)};
    \addplot[PL] coordinates {(1, 0.6144) (5, 0.5681999999999999) (10, 0.6008) (15, 0.6739999999999999) (20, 0.5935999999999999) (25, 0.5367999999999999) (30, 0.5664)};
    \addplot[TS] coordinates {(1, 0.5988000000000001) (5, 0.5900000000000001) (10, 0.56) (15, 0.5802) (20, 0.6268) (25, 0.5988) (30, 0.6184000000000001)};
    \addplot[Trend] coordinates {(1, 0.5964) (5, 0.5808) (10, 0.616) (15, 0.5472) (20, 0.5576000000000001) (25, 0.6518) (30, 0.6216)};
    \addplot[Seasonality] coordinates {(1, 0.6536) (5, 0.5827999999999999) (10, 0.643) (15, 0.5956) (20, 0.6683999999999999) (25, 0.54) (30, 0.5952)};
    \addplot[MASE] coordinates {(1, 0.6) (5, 0.6414) (10, 0.5448000000000001) (15, 0.5880000000000001) (20, 0.5904) (25, 0.569) (30, 0.6172)};
    \addplot[MSE] coordinates {(1, 0.6376) (5, 0.6704) (10, 0.5684) (15, 0.6638000000000001) (20, 0.5748) (25, 0.588) (30, 0.5504)};
    \addplot[MSEMASE] coordinates {(1, 0.5608) (5, 0.6574) (10, 0.6064) (15, 0.6032) (20, 0.6517999999999999) (25, 0.6262) (30, 0.614)};
    \nextgroupplot[xlabel={Horizon}, ylabel={{AUC-ROC}}, title={\textbf{LSTM - ECG}}]
    \addplot[uppbound] coordinates {(1, 0.5951353086419753) (5, 0.5744241975308642) (10, 0.560531522633745) (15, 0.5662735802469135) (20, 0.5684495473251029) (25, 0.5766762139917695) (30, 0.5887107818930041)};    
    \addplot[All] coordinates {(1, 0.5871999999999999) (5, 0.6084) (10, 0.6312) (15, 0.5728000000000001) (20, 0.6129) (25, 0.5880000000000001) (30, 0.6214999999999999)};
    \addplot[PL] coordinates {(1, 0.5828) (5, 0.6584) (10, 0.6706) (15, 0.5) (20, 0.5634999999999999) (25, 0.6077999999999999) (30, 0.6068)};
    \addplot[TS] coordinates {(1, 0.6160000000000001) (5, 0.62) (10, 0.5968) (15, 0.6836) (20, 0.5611999999999999) (25, 0.5808) (30, 0.589)};
    \addplot[Trend] coordinates {(1, 0.632) (5, 0.596) (10, 0.5592) (15, 0.5790000000000001) (20, 0.5732) (25, 0.6302000000000001) (30, 0.6284000000000001)};
    \addplot[Seasonality] coordinates {(1, 0.5700000000000001) (5, 0.59) (10, 0.576) (15, 0.5584) (20, 0.6235999999999999) (25, 0.5879) (30, 0.6189)};
    \addplot[MASE] coordinates {(1, 0.5821999999999999) (5, 0.5988) (10, 0.6288) (15, 0.591) (20, 0.6241000000000001) (25, 0.6252) (30, 0.5865)};
    \addplot[MSE] coordinates {(1, 0.5736000000000001) (5, 0.5928) (10, 0.6228) (15, 0.5728) (20, 0.6219999999999999) (25, 0.6343000000000001) (30, 0.579)};
    \addplot[MSEMASE] coordinates {(1, 0.6104) (5, 0.617) (10, 0.5798) (15, 0.54) (20, 0.6008) (25, 0.6112) (30, 0.6002000000000001)};
\addlegendentry{Benchmark Limit}
\addlegendentry{PV, T and S}
\addlegendentry{PV and T}
\addlegendentry{PV, MSE and MASE}
\addlegendentry{PV}
\addlegendentry{PV and MASE}
\addlegendentry{PV and MSE}
\addlegendentry{PV and S}
\addlegendentry{PV, T, S, MSE and MASE}
  \end{groupplot}
\end{tikzpicture}
\end{adjustbox}
\caption{AUC-ROC Performance Across Different Models on the EEG and ECG datasets. The Predicted Values feature was added to each feature combination}
\label{fig:ecg_pl}
\end{figure*}

We examined the effects of incorporating Predicted Values (PV) as a feature in the attack model. The Predicted Values were added to every feature combination and tested independently. The results on the EEG and ECG datasets are illustrated in Figure \ref{fig:ecg_pl}, with some models omitted due to space limitations.
The black curve, labeled "Benchmark Limit", represents the upper bound of the AUC-ROC values for the feature combinations without Predicted Values, as presented in the previous section. 

The analysis of the results did not reveal a consistent trend across horizons and datasets. For the EEG dataset, adding the Predicted Values consistently under-performed compared to the feature combinations without them.
In contrast, the results for the ECG dataset were more varied. Although the addition of the Predicted Values feature often surpassed the Benchmark Limit across different models and horizons, the results varied widely and no feature combination consistently exceeded the Benchmark Limit across all horizons.

Overall, the results indicate that while the Predicted Values feature can in some cases improve the performance of the attack model, its impact is not straightforward and depends on the specific model, dataset and horizon. The figures highlight the complexity and variability of incorporating Predicted Values in this context.

\subsection{True Positive Rate at 1\% False Positive Rate Results}
In addition to AUC-ROC, we measured the attack's True Positive Rate at a False Positive Rate of 1\% for a fixed horizon of 5. The results are presented in Tables \ref{tab:eeg_tpr@fpr} and \ref{tab:ecg_tpr@fpr}. 

For the EEG dataset, using the Seasonality (S) feature with the PatchTST model resulted in a 9x increase in TPR compared to using the MSE feature alone. Significant improvements were observed across the NFT, TimesNet, and LSTM models when using the Trend (T) feature, with gains ranging from 2.1x to 4x over the MSE-only baseline. The TCN model demonstrated a 2.1x improvement with the combination of Trend and Seasonality, while DLinear showed similar gains with both combined and standalone Seasonality features.
In the ECG dataset, the Trend feature and its combination with Seasonality in the NFT model led to a 3x improvement in TPR compared to the MSE-only baseline. The Seasonality feature led to notable improvements for the TimesNet and PatchTST models, with up to a 12x increase.

\begin{table}[h]
\centering
    \caption{Average TPR at 1\% FPR for EEG Dataset at Horizon 5}
    \label{tab:eeg_tpr@fpr}
    \begin{tabular}{lcccccccc}
        \toprule
        Model &  MSE & MASE & MSE and & T & S & T and & T, S, MASE, & PV \\
              &      &      &  MASE   &   &   & S     & MSE, and PV &     \\
        \midrule
        DLinear & 0.03 & 0.02 & 0.04 & 0.06 & \textbf{0.07} & \textbf{0.07} & 0.06 & 0.06 \\
        TCN & 0.23 & 0.04 & 0.23 & 0.40 & 0.42 &\textbf{ 0.43} & 0.35 & 0.31 \\
        LSTM & 0.18 & 0.10 & 0.20 & \textbf{0.44} & 0.39 & 0.38 & 0.39 & 0.27 \\
        NFT & 0.20 & 0.06 & 0.21 & \textbf{0.42} & 0.37 & 0.41 & 0.38 & 0.30 \\
        TimesNet & 0.02 & 0.01 & 0.03 & \textbf{0.08 }& 0.06 & 0.05 & 0.06 & 0.06 \\
        PatchTST & 0.07 & 0.02 & 0.06 & 0.28 & \textbf{0.63} & 0.39 & 0.35 & 0.13 \\
        \bottomrule
    \end{tabular}
\end{table}

\begin{table}[h]
\centering
    \caption{Average TPR at 1\% FPR for ECG Dataset at Horizon 5}
    \label{tab:ecg_tpr@fpr}
    \begin{tabular}{lcccccccc}
        \toprule
        Model &  MSE & MASE & MSE and & T & S & T and & T, S, MASE, & PV \\
              &      &      &  MASE   &   &   & S     & MSE, and PV &     \\
        \midrule
        DLinear & 0.01 & 0.01 & 0.02 & \textbf{0.03} & \textbf{0.03} & \textbf{0.03} & \textbf{0.03} & \textbf{0.03} \\
        TCN & 0.01 & 0.01 & 0.01 & \textbf{0.02} & 0\textbf{.02} & \textbf{0.02} & \textbf{0.02} & \textbf{0.02} \\
        LSTM & 0.02 & 0.02 & 0.01 & 0.02 & 0.02 & 0.01 & 0.02 & \textbf{0.04} \\
        NFT & 0.01 & 0.02 & 0.02 & \textbf{0.03} & 0.02 & \textbf{0.03} & 0.02 & 0.02 \\
        TimesNet & 0.03 & 0.05 & 0.03 & 0.14 & \textbf{0.36} & 0.35 & 0.36 & 0.03 \\
        PatchTST & 0.34 & 0.03 & 0.32 & 0.52 & \textbf{0.99} & 0.51 & 0.66 & 0.03 \\
        \bottomrule
    \end{tabular}
    
\end{table}

\section{Conclusion and Future Work}
\label{conc}
This paper is the first to explore membership inference attacks on numeric time-series machine-learning models. Our primary contribution is the introduction of two novel features, Trend and Seasonality, derived from low-degree polynomial fitting and Multivariate Fourier Transform respectively. These features enhance MIA models by leveraging the inherent characteristics of time-series data, improving the identification of member samples.

Various state-of-the-art forecasting models incorporate those components into their design. Consequently, when targeting a time-series prediction model, there is a significant likelihood that the model will precisely estimate the series' Seasonality and Trend of its training data, providing a strategic advantage in MIAs. 

The effectiveness of these features was tested on six diverse models using two medical datasets. The Trend and Seasonality features showed superior accuracy, with a 3\% to 26\% improvement in attack AUC-ROC scores over traditional features across various horizons.

We plan to explore additional MIA scenarios, including models pre-trained on many patients and fine-tuned for home monitoring of a specific patient, to see if such an attack can expose the original training data. Additionally, we aim to investigate user-level attacks that exploit the fact that multiple samples in the training set belong to the same person.

\section*{Acknowledgment}
This work was performed as part of the NEMECYS project, which is co-funded by the European Union under grant agreement ID 101094323, by UK Research and Innovation (UKRI) under the UK government’s Horizon Europe funding guarantee grant numbers 10065802, 10050933, and 10061304, and by the Swiss State Secretariat for Education, Research and Innovation (SERI). 

\bibliography{references}


\end{document}